\documentclass[letterpaper, 10 pt, conference]{ieeeconf}  

\IEEEoverridecommandlockouts
\usepackage{commands}

\title{\LARGE \bf
	PLK-Calib: Single-shot and Target-less LiDAR-Camera Extrinsic Calibration using Pl\"ucker Lines
}

\author{Yanyu Zhang, Jie Xu, Wei Ren% <-this % stops a space
\thanks{Y. Zhang, J. Xu, and W. Ren are with the Department of Electrical and Computer Engineering, University of California, Riverside, CA, 92521, USA. Email: \{yzhan831, jxu150, weiren\}@ucr.edu.}%
}

\begin{document}

\maketitle
\thispagestyle{empty}
\pagestyle{empty}

%%%%%%%%%%%%%%%%%%%%%%%%%%%%%%%%%%%%%%%%%%%%%%%%%%%%%%%%%%%%%%%%%%%%%%%%%%%%%%%%
\begin{abstract}
Accurate LiDAR-Camera (LC) calibration is challenging but crucial for autonomous systems and robotics. In this paper, we propose two single-shot and target-less algorithms to estimate the calibration parameters between LiDAR and camera using line features. The first algorithm constructs line-to-line constraints by defining points-to-line projection errors and minimizes the projection error. The second algorithm (PLK-Calib) utilizes the co-perpendicular and co-parallel geometric properties of lines in Pl\"ucker (PLK) coordinate, and decouples the rotation and translation into two constraints, enabling more accurate estimates. Our degenerate analysis and Monte Carlo simulation indicate that three nonparallel line pairs are the minimal requirements to estimate the extrinsic parameters. Furthermore, we collect an LC calibration dataset with varying extrinsic under three different scenarios and use it to evaluate the performance of our proposed algorithms.
\end{abstract}

\section{INTRODUCTION}\label{sec:Introduction}
LiDAR and cameras are two primary sensors utilized in autonomous driving (AD) and simultaneous localization and mapping (SLAM). To fuse the data from different sensors, precise spatial and temporal calibrations are required \cite{Zhang2024}. Spatial calibration typically involves global registration followed by local refinement. In this paper, we address the local refinement of extrinsic parameters between LiDAR and camera, focusing on optimizing the relative transformation between two coordinate systems, which includes both rotation and translation.

To obtain a high-precision extrinsic matrix, robust feature extraction and matching are essential for establishing correspondences between the LiDAR point cloud and the camera image. In the LiDAR-Camera (LC) calibration literature, points \cite{Huang2024, Petek2024, Koide2023, Boretti2022}, lines \cite{Yuan2021, Bai2021, Zhang2025}, and planes \cite{Huang2024_2, Zhaotong2023} are commonly employed for this purpose. Compared to lines or planes, points are simple but involve fewer constraints, and point extraction from the point cloud can be less accurate, particularly with a sparse LiDAR such as Ouster and Velodyne. Additionally, points are susceptible to noise, especially in environments with poor visibility or complex textures. In contrast, line and plane features can mitigate the impact of noise and provide robust geometric constraints. However, detecting and accurately modeling planes can be challenging, particularly in environments with intricate textures or irregular surfaces. Furthermore, plane-based calibration may not be suitable for all types of scenes or objects. Considering the advantages and limitations of all three features, we focus on using line features exclusively to estimate the extrinsic matrix between LiDAR and camera.

After extracting and matching the line pairs between point clouds and images, the 2D-to-3D line constraints are constructed, and the rotation and translation between the LiDAR and camera are estimated simultaneously \cite{Yuan2021, Bai2021}. However, in most of the LC calibration literature, estimating the translation is more challenging than the rotation, with translation errors typically being much higher than rotation errors \cite{Huang2024, Koide2023, Yuan2021, Bai2021}. There are two main reasons: (a) translation is more sensitive to measurement errors, even a small orientation error can impact translation accuracy significantly, (b) the nature of scenarios, such as those involving 3D line or plane patterns, makes accurately translating coordinates more difficult than aligning them rotationally.

To address the aforementioned challenges, we propose two single-shot and target-less LC calibration algorithms using line features. We assume that line features are present in the current scene, and the camera and LiDAR have overlapping fields of view. Given an initial guess, the first algorithm constructs line-to-line constraints by projecting the 3D lines from the LiDAR frame onto the image frame, minimizing the projection error. The second algorithm utilizes the co-perpendicular and co-parallel constraints in the Pl\"ucker (PLK) coordinate, and optimizes the rotation first. Once the rotation is determined, the translation between LiDAR and camera frame can be estimated by solving a least squares problem. The main contributions of our work include:
\vspace{-1pt}
\begin{itemize}
  \item We propose two single-shot and target-less LC calibration algorithms using Pl\"ucker lines. The first algorithm, which will be used \textit{as a baseline}, minimizes line-to-line projection errors. The second algorithm (PLK-Calib) minimizes the co-perpendicular and co-parallel constraint errors. Unlike traditional methods, PLK-Calib decouples the rotation and translation constraint, leading to a more accurate estimate.
  \item We analyze the degenerate cases of both algorithms and verify through Monte Carlo simulations, demonstrating that three nonparallel line pairs are the minimal requirement for LC extrinsic optimization.
  \item We collect an LC calibration dataset across multiple scenarios, which could be beneficial to researchers in the calibration field. Unlike all previous LC datasets, our dataset allows for varying extrinsic parameters through the use of a mechanical arm. The robustness of the algorithms can be experimentally validated by calibrating different extrinsic parameters in the same scene. We then evaluate our proposed algorithms using these data sequences and compare them with the \textit{state-of-the-art} works.
\end{itemize}

% In the rest of this report, we first summarize the related work in Sec.~\ref{sec:RelatedWork}. The details of our algorithm are introduced in Sec.~\ref{sec:Methodology}. Then, a degenerate analysis through simulation and real-world experiments are presented in Sec.~\ref{sec:Simulations} and Sec.~\ref{sec:Experiments}, respectively. Finally, we summarize and outline future work in Sec.~\ref{sec:Conclusions}.

\section{RELATED WORK}\label{sec:RelatedWork}
Based on various perspectives, LC extrinsic calibration techniques can be categorized into: 1) target-based or target-less, 2) multi-shot or single-shot.

\subsection{Target-based Calibration}\label{sec:TargetBasedCalibration}
Target-based calibration refers to algorithms that utilize a known target, such as a checkerboard \cite{calibration_target}, AprilGrid \cite{Olson2011}, or a custom board. The calibration process begins by identifying the features of the target in both the point cloud and the image. Once the correspondences are established, the 3D points are projected onto the image. The LC transformation is then estimated by solving the perspective-n-point (PnP) problem. A significant challenge is designing a calibration target that can be reliably and accurately detected by both LiDAR and camera. In \cite{Ankit2017}, an algorithm is proposed to find 3D-to-3D point correspondences between the point cloud and the image, solving for the extrinsic matrix using a closed-form solution. To enhance the robustness of the calibration target, a novel custom board combining AprilGrid and CircleGrid \cite{calibration_target} patterns is introduced in \cite{Verma2019}. However, this specific board is complex for users to fabricate. An efficient checkerboard-based method is proposed in \cite{Huang2020}, which optimizes the normal direction and corner position of the checkerboard plane. Further improvements are offered in \cite{Beltrán2022} by incorporating edge constraints of the checkerboard and additional point constraints located on the checkerboard.

\subsection{Target-less Calibration}\label{sec:TargetLessCalibration}
Recently, target-less calibration has gained significant attention due to its flexibility. A pixel-level accuracy target-less LC extrinsic calibration algorithm is proposed in \cite{Yuan2021}, but it requires a high-resolution LiDAR like Livox AVIA. Additionally, a general, single-shot, target-less calibration algorithm has been introduced in \cite{Koide2023}. However, achieving a dense single-shot point cloud with this method necessitates either the use of a dense LiDAR or the integration of multiple LiDAR scans using the CT-ICP \cite{Dellenbach2022} algorithm. In addition to the Levenberg-Marquardt (LM)-based solution, there are also some learning-based target-less approaches. Earlier learning-based works use a convolutional neural network (CNN) to optimize the extrinsic matrix by inputting a single point cloud and an image \cite{Schneider2017, Iyer2018}. More recently, a transformer-based LC calibration framework has been proposed in \cite{Yuxuan2024}. By combining the advantages of rich feature extraction from learning-based methods and the accurate estimation of the extrinsic matrix from LM-based methods, a hybrid target-less, cross-modality algorithm has been proposed by \cite{Huang2024}.

\subsection{Multi-shot Calibration}\label{sec:MultiShotCalibration}
Multi-shot calibration involves the use of multiple continuous images and point clouds as input for calibration. Typically, multi-shot calibration is associated with motion estimation and can be integrated into online SLAM calibration processes. In \cite{Petek2024}, they separately estimate visual and LiDAR odometry, subsequently aligning the trajectories of the two sensors to achieve coarse registration. Given a good initial guess, an online LC calibration algorithm is proposed for Livox LiDAR in \cite{Zhu2021}. Furthermore, to monitor the calibration quality, an online calibration drift tracking system is provided in \cite{Moravec2024}.

\subsection{Single-shot Calibration}\label{sec:SingleShotCalibration}
Compared to multi-shot calibration, single-shot LC calibration is more challenging due to the limited features and the less accurate correspondences. To enhance feature extraction reliability, many studies employ learning-based methods as the frontend \cite{Schneider2017, Iyer2018, Zhaotong2023, Yuxuan2024, Huang2024, Huang2024_2}. Transformer-based methods are highlighted in \cite{Yuxuan2024} and \cite{Huang2024}. In Calib-Anything \cite{Zhaotong2023}, the Segment Anything (SAM) \cite{Kirillov2023} is used to extract segments from the image. The LiDAR point cloud is then projected onto the image frame, and minimizes the distance between projected points and objects. Benefiting from the dense Livox LiDAR, SAM is employed on both the LiDAR intensity projection (LIP) and RGB images \cite{Huang2024_2}, minimizing the object-to-object distance.

In this paper, we address the most challenging scenarios involving target-less and single-shot inputs. Additionally, our approach does not require the use of a dense LiDAR and does not rely on any specific environmental setup.

\section{METHODOLOGY}\label{sec:Methodology}
The goal of the PLK-Calib is to estimate the calibration extrinsic parameter $\mathbf{T}_{L}^{C} \triangleq \,$\{${ }_{L}^{C}\mathbf{R}, { }^{C}\mathbf{P}_{L}$\} between the LiDAR frame \{$L$\} and the camera frame \{$C$\} using line features given a roughly initial guess. We assume that the current scene contains at least three lines, which is a common setting in most human-made environments. Besides, we require the camera and LiDAR to have overlapping fields of view. With these prerequisites, we propose two LC extrinsic calibration algorithms using PLK lines. The first algorithm directly minimizes the projection error. The second algorithm (PLK-Calib) leverages the special co-perpendicular and co-parallel constraints of PLK lines to decouple the rotation and translation components from the extrinsic matrix. Specifically, the rotational component is estimated first and then used as a known parameter when optimizing the translation.

\subsection{Method I: Projection Error Minimization}
\label{sec:method1}
\begin{figure}[t]
    \centering
    \subfigure[]{\label{fig:cp_line}\definecolor{lightblue}{rgb}{0.212, 0.451, 0.792}

\begin{tikzpicture}[line width=1pt,scale=0.5]
\draw[->] (0,0,0) -- (0,0,4) node[anchor=south, xshift=-3]{$x$};
\draw[->] (0,0,0) -- (4,0,0) node[anchor=north east]{$y$};
\draw[->] (0,0,0) -- (0,4,0) node[anchor=north east]{$z$};
\node at (0,0,0) [anchor=south east, yshift=-6] {$L$};

\draw[red] (0.85,3.13,0.9) -- (4,0.5,3);
\draw[red, dashed] (0,0,0) -- (0.85,3.13,0.9);
\node at (0.85,3.13,0.9) [anchor=south, xshift=3]{${}^{L}\mathbf{p}_{f1}$};
\draw[red, dashed] (0,0,0) -- (4,0.5,3); 
\node at (4,0.5,3) [anchor=north]{${}^{L}\mathbf{p}_{f2}$};
\fill[red] (0.85,3.13,0.9) circle (4pt);
\fill[red] (4,0.5,3) circle (4pt);

\draw[lightblue, dashed] (0,0,0) -- (2.02, 1.81, 1.34);
\node at (2.02, 1.81, 1.34) [anchor=north, xshift=-7, yshift=-5]{$d_l$};
\draw[black] (1.62, 1.41, 0.94) -- (1.95, 1.22, 1.2) -- (2.32, 1.56, 1.54);

\draw[lightblue, dashed, ->] (2.02, 1.81, 1.34) -- (1.12, 2.56, 0.74);
\node at (1.12, 2.56, 0.74) [anchor=north, xshift=-1, yshift=-2]{${}^{L}\mathbf{v}_l$};
\draw[lightblue, dashed, ->] (2.02, 1.81, 1.34) -- (2.52, 2.56, 0.94);
\node at (2.52, 2.56, 0.94) [anchor=north west, xshift=-2]{${}^{L}\mathbf{n}_l$};

\end{tikzpicture}}
    \subfigure[]{\label{fig:perpendicular}\definecolor{lightblue}{rgb}{0.212, 0.451, 0.792}
\definecolor{darkgreen}{rgb}{0.0, 0.5, 0.0}
\definecolor{darkred}{rgb}{0.55, 0.0, 0.0}

\begin{tikzpicture}[line width=1pt,scale=0.5]
% camera part
\draw[->] (0,0,0) -- (1,0,0) node[anchor=north west, xshift=-5]{$x$};
\draw[->] (0,0,0) -- (0,-1,0) node[anchor=north east, yshift=5]{$y$};
\draw[->] (0,0,0) -- (0,0,-1.8) node[anchor=south east, yshift=-2]{$z$};
\node at (0,0,0) [anchor=east] {$C$};

\draw[red] (2.5,1,0) -- (2.5,4,0);
\draw[lightblue, dashed, ->] (2.5,2,0) -- (3.3,2.8,0);
\node at (2.5,2,0) [anchor=east, xshift=2, yshift=3] {${}^{C}\mathbf{n}_{l}^\prime$};
\draw[red, dashed] (0,0,0) -- (2.5,1,0);
\draw[red, dashed] (0,0,0) -- (2.5,4,0);
\fill[red] (2.5,1,0) circle (3pt);
\fill[red] (2.5,4,0) circle (3pt);
\draw[darkred] (1.5,0.6,0) -- (1.25,2,0);
\node at (1.5,0.6,0) [anchor=south west, xshift=-3, yshift=2] {$l^\prime$};

\draw[lightblue] (0.7,1.4,0) -- (0.7,3,0);
\draw[lightblue] (2,-0.2,0) -- (2,1.4,0);
\draw[lightblue] (0.7,3,0) -- (2,1.4,0);
\draw[lightblue] (2,-0.2,0) -- (0.7,1.4,0);

\draw[->] (5,0,0) -- (4.3,0.7,0) node[anchor=south]{$x$};
\draw[->] (5,0,0) -- (4,0,0) node[anchor=north]{$y$};
\draw[->] (5,0,0) -- (5,1,0) node[anchor=west]{$z$};
\node at (5,0,0) [anchor=west] {$L$};

% LiDAR part
\draw[darkgreen] (2.8,1,0) -- (2.8,4,0);
\draw[lightblue, dashed, ->] (2.8,4,0) -- (2.8,4.7,0);
\node at (2.8,4.5,0) [anchor=west, xshift=-3] {${}^{L}\mathbf{v}_{l}$};
\draw[lightblue, dashed, ->] (3.5,1.5,0) -- (4.5,2.5,0);
\node at (4.5,2.5,0) [anchor=south] {${}^{L}\mathbf{n}_{l}$};

\fill[darkgreen] (2.8,1,0) circle (3pt);
% \fill[darkgreen] (2.8,2,0) circle (3pt);
% \fill[darkgreen] (2.8,3,0) circle (3pt);
\fill[darkgreen] (2.8,4,0) circle (3pt);

\draw[darkgreen, dashed] (5,0,0) -- (2.8,1,0);
% \draw[darkgreen, dashed] (5,0,0) -- (2.8,2,0);
% \draw[darkgreen, dashed] (5,0,0) -- (2.8,3,0);
\draw[darkgreen, dashed] (5,0,0) -- (2.8,4,0);

\end{tikzpicture}}
    \caption{Geometric description of PLK line and constraints. (a) The red solid line represents a 3D line in the LiDAR coordinate with two endpoints ${}^{L}\mathbf{p}_{f1}$ and ${}^{L}\mathbf{p}_{f2}$. The vector ${}^{L}\mathbf{v}_l$ denotes the line direction, while ${}^{L}\mathbf{n}_l$ is a vector that is perpendicular to the line plane, which consists of two endpoints and the origin. (b) The green solid line corresponds to the same line depicted in (a), and $l^\prime$ represents the 2D line measurements in the camera frame. The red line is back-projected from $l^\prime$. Due to camera measurement noise, there are some rotation and translation discrepancies between red and green solid lines, although they indicate the same line. Our algorithm minimizes this residual by optimizing the extrinsic parameters.}
    \vspace*{-10pt}
    \label{fig:constraints}
\end{figure}
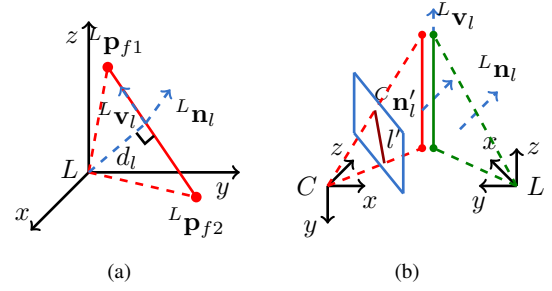

For a 3D line, we adopt the PLK lines in our previous work \cite{Zhang2023}, which represents the 3D line by two $3\times1$ vectors. As shown in Fig.~\ref{fig:cp_line}, given two points ${}^{L}\mathbf{p}_{f1}$ and ${}^{L}\mathbf{p}_{f2}$ on a 3D line in LiDAR frame, the line under PLK coordinate can be expressed by:
\begin{equation}
	\begin{aligned}
		\left[\begin{array}{l}
			{}^{L}\mathbf{n}_{l} \\
			{}^{L}\mathbf{v}_{l}
		\end{array}\right]=
        \left[\begin{array}{l}
			\left\lfloor{}^{L}\mathbf{p}_{f1}\times\right\rfloor {}^{L}\mathbf{p}_{f2} \vspace*{2pt}\\
			{}^{L}\mathbf{p}_{f2}-{}^{L}\mathbf{p}_{f1}
		\end{array}\right],
		\label{eq:pluker_line}
	\end{aligned}
\end{equation}
where line plane is the plane consisting of two endpoints and the origin, and $\lfloor\,\cdot\,\times\rfloor$ denotes skew-symmetric matrix \cite{Trawny2005}. The plane consists of two endpoints and the origin is the line plane. ${}^{L}\mathbf{n}_{l}$ denotes the normal direction of the line plane and ${}^{L}\mathbf{v}_{l}$ is the line direction.

% Then, the CPL can be expressed as:
% \begin{equation}
% 	\begin{aligned}
% 		{}^{L} \mathbf{x}_l = d_l \bar{q_l} = \left[\mathbf{q}_l^{\top} \quad q_l \right]^{\top},
% 		\label{eq:cp_line}
% 	\end{aligned}
% \end{equation}
% where the distance can be computed as $d_{l}=\left\|{}^{L}\mathbf{n}_{l}\right\| /\left\|{}^{L}\mathbf{v}_{l}\right\|$. The unit quaternion $\bar{q_l}$ can be transformed from $\mathbf{R}\left(\bar{q}_{l}\right)=\left[{}^{L}\mathbf{n}_{e} \ {}^{L}\mathbf{v}_{e} \ \lfloor{}^{L}\mathbf{n}_{e}\times\rfloor {}^{L}\mathbf{v}_{e} \right]$, where ${}^{L}\mathbf{n}_{e}$ and ${}^{L}\mathbf{v}_{e}$ are the unit vectors of ${}^{L}\mathbf{n}_{l}$ and ${}^{L}\mathbf{v}_{l}$.

Given an initial guess of transformation between the LiDAR frame and the camera frame, a PLK line can be transformed from the LiDAR frame to the camera frame as:
\begin{equation}
    \begin{aligned}
    	{ }^{C} \mathbf{L}=
    	\left[\begin{array}{cc}
    		{ }_{L}^{C}\mathbf{R} & \lfloor { }^{C}\mathbf{P}_{L}\times\rfloor { }_{L}^{C}\mathbf{R}\\
    		\mathbf{0}_3 & { }_{L}^{C}\mathbf{R}
    	\end{array}\right]
    	{ }^{L} \mathbf{L}
        \label{eq:line_transformation}
    \end{aligned}
\end{equation}
where ${ }^{L} \mathbf{L} = \left[{ }^{L} \mathbf{n}_{l}^\top \; { }^{L} \mathbf{v}_{e}^\top\right]^\top = \left[{ }^{L}d_l { }^{L} \mathbf{n}_{e}^\top \; { }^{L} \mathbf{v}_{e}^\top\right]^\top$ and ${ }^{C} \mathbf{L} = \left[{ }^{C} \mathbf{n}_{l}^\top \; { }^{C} \mathbf{v}_{e}^\top\right]^\top = \left[{ }^{C}d_l { }^{C} \mathbf{n}_{e}^\top \; { }^{C} \mathbf{v}_{e}^\top\right]^\top$ are the PLK line in the LiDAR frame and camera frame, respectively. Then, we adopt the projective line measurement model \cite{Adrien2005} and the 3D line can be projected to the camera plane as:
\begin{equation}
    \begin{aligned}
        \left[\begin{array}{c}
            l_1 \\ l_2 \\ l_3 \end{array}\right] &=
        \underbrace{\left[\begin{array}{ccc}
            f_{v} & 0 & 0\\
            0 & f_{u} & 0\\
            -f_{v} c_{u} & -f_{u} c_{v} & f_{u} f_{v}
        \end{array}\right.}_\mathbf{K} \left.
        \begin{array}{|ccc}
            0 & 0 & 0\\
            0 & 0 & 0\\
            0 & 0 & 0
        \end{array}\right] {}^{C} \mathbf{L} \\
        &\triangleq \mathbf{K} \, { }^{C}\mathbf{n}_{l},
        \label{eq:line_projection}
    \end{aligned}
\end{equation}
where $f_{u}$, $f_{v}$, $c_{u}$, $c_{v}$ are the camera intrinsic parameters. The 2D line residual between the projected line and two detected line endpoints, $\mathbf{x}_{s} = \left[u_{s} \ v_{s} \ 1\right]^{\top}$ and $\mathbf{x}_{e} = \left[u_{e} \ v_{e} \ 1\right]^{\top}$ can be expressed as:
\begin{equation}
	\begin{aligned}
		\mathbf{r}=
        \left[\begin{array}{ll}
			\frac{\mathbf{x}_{s}^{\top} \mathbf{l}}{\sqrt{l_{1}^{2}+l_{2}^{2}}} & \frac{\mathbf{x}_{e}^{\top} \mathbf{l}}{\sqrt{l_{1}^{2}+l_{2}^{2}}}
		\end{array}\right]^{\top},
        \label{eq:line_residual}
    \end{aligned}
\end{equation}
where $\mathbf{l} = \left[l_1 \ l_2 \ l_3\right]^{\top}$ denotes the 2D line on the image.

By adding $N$ 3D-to-2D line residuals, we minimize the loss function by optimizing the LC extrinsic parameters as: 
\begin{equation}
    \begin{aligned}
        \underset{{}_{L}^{C}\mathbf{\tilde{\theta}}, \, {}^{C}\mathbf{\tilde{P}}_{L}}{\text{minimize}}
        \sum_{n=1}^{N} \mathbf{r}_{n}.
        \label{eq:line_minimal}
    \end{aligned}
\end{equation}
To get the analytical solution, the derivation of Jacobians can be expressed as:
\begin{equation}
    \begin{aligned}
        \frac{\partial \mathbf{r}}{\partial {}_{L}^{C}\mathbf{\tilde{\theta}}} = \frac{\partial \mathbf{r}}{\partial \mathbf{l}} \frac{\partial \mathbf{l}}{\partial {}^{C}\mathbf{L}} \frac{\partial {}^{C}\mathbf{L}}{\partial {}_{L}^{C}\mathbf{\tilde{\theta}}}, \;\text{and}\;
        \frac{\partial \mathbf{r}}{\partial { }^{C}\mathbf{\tilde{P}}_{L}} = \frac{\partial \mathbf{r}}{\partial \mathbf{l}} \frac{\partial \mathbf{l}}{\partial {}^{C}\mathbf{L}} \frac{\partial {}^{C}\mathbf{L}}{\partial { }^{C}\mathbf{\tilde{P}}_{L}},
        \label{eq:line_Jacobians}
    \end{aligned}
\end{equation}
where
\begin{equation}
    \begin{aligned}
        \frac{\partial \mathbf{r}}{\partial \mathbf{l}} = 
        \frac{1}{\sqrt{l_1^2+l_2^2}}\left[\begin{array}{ccc}
        u_s-\frac{l_1 \mathbf{x}_{s}^{\top}\mathbf{l}}{l_1^2+l_2^2} & v_s-\frac{l_2 \mathbf{x}_{s}^{\top}\mathbf{l}}{l_1^2+l_2^2} & 1 \vspace*{2pt}\\ 
        u_e-\frac{l_1 \mathbf{x}_{e}^{\top}\mathbf{l}}{l_1^2+l_2^2} & v_e-\frac{l_2 \mathbf{x}_{e}^{\top}\mathbf{l}}{l_1^2+l_2^2} & 1
        \end{array}\right],
        \label{eq:line_Jacobians1}
        \nonumber
    \end{aligned}
\end{equation}
\begin{equation}
    \begin{aligned}
        \frac{\partial \mathbf{l}}{\partial {}^{C}\mathbf{L}} = \frac{\partial \mathbf{l}}{\left[{ }^{C} \mathbf{n}_{l}^\top \ { }^{C} \mathbf{v}_{e}^\top\right]^\top} = 
        \left[\begin{array}{cc}
        \mathbf{K} & \mathbf{0}_3
        \end{array}\right],
        \label{eq:line_Jacobians2}
        \nonumber
    \end{aligned}
\end{equation}
\begin{equation}
    \begin{aligned}
        \frac{\partial {}^{C}\mathbf{L}}{\partial {}_{L}^{C}\mathbf{\tilde{\theta}}} = 
        \left[\begin{array}{c}
        {}^{L}d_l\lfloor{}_{L}^{C}\mathbf{R} {}^{L} \mathbf{n}_{e}\times\rfloor + \lfloor{ }^{C}\mathbf{P}_{L}\times\rfloor \lfloor{}_{L}^{C}\mathbf{R} {}^{L} \mathbf{v}_{e}\times\rfloor \vspace*{2pt}\\
        \lfloor{}_{L}^{C}\mathbf{R} {}^{L} \mathbf{v}_{e}\times\rfloor
        \end{array}\right],
        \label{eq:line_Jacobians3}
        \nonumber
    \end{aligned}
\end{equation}
and
\begin{equation}
    \begin{aligned}
        \frac{\partial {}^{C}\mathbf{L}}{\partial { }^{C}\mathbf{\tilde{P}}_{L}} = 
        \left[\begin{array}{c}
        -\lfloor{}_{L}^{C}\mathbf{R} {}^{L} \mathbf{v}_{e}\times\rfloor\\
        \mathbf{0}_3
        \end{array}\right].
        \label{eq:line_Jacobians4}
        \nonumber
    \end{aligned}
\end{equation}

\subsection{Method II (PLK-Calib): Co-Perpendicular Error and Co-Parallel Error Minimization}
\label{sec:method2}
Given a 2D line in the image frame, expressed as $l^\prime = [l_1^\prime,l_2^\prime,l_3^\prime]^\top$, the corresponding 3D line in the camera PLK coordinate, denoted by ${ }^{C} \mathbf{L}^\prime = [{{}^{C}\mathbf{n}_{l}^\prime}^\top, {{}^{C}\mathbf{v}_{l}^\prime}^\top]^\top$, can be partially determined through back projection according to \eqref{eq:line_projection}. Specifically, we can obtain the back-projected 3D line direction as:
\begin{equation}
    \begin{aligned}
        {}^{C}\mathbf{n}_{l}^\prime =(\mathbf{K})^{-1}\left[\begin{array}{c} l_1^\prime \\ l_2^\prime \\ l_3^\prime \end{array}\right].
        \label{eq:back_projection}
    \end{aligned}
\end{equation}
Recall that the actual line correspondence represented in the LiDAR PLK coordinate is ${ }^{L} \mathbf{L} = [{{}^{L}\mathbf{n}_{l}}^\top, {{}^{L}\mathbf{v}_{l}}^\top]^\top$. We can rewrite the line transformation from LiDAR to camera frame in \eqref{eq:line_transformation} as:
\begin{equation}
    \begin{aligned}
        {}^{C}\mathbf{n}_{l} &= { }_{L}^{C}\mathbf{R} {}^{L}\mathbf{n}_{l} + \lfloor { }^{C}\mathbf{P}_{L}\times\rfloor { }_{L}^{C}\mathbf{R}  {}^{L}\mathbf{v}_{l}, \\
        {}^{C}\mathbf{v}_{l} &= { }_{L}^{C}\mathbf{R}  {}^{L}\mathbf{v}_{l}.
        \label{eq:line_transformation2}
    \end{aligned}
\end{equation}
Because the transformed ${}^{C} \mathbf{x}_l \triangleq [{}^{C}\mathbf{n}_{l}^\top,{}^{C}\mathbf{v}_{l}^\top]^\top$ and ${}^{C} \mathbf{x}_l^\prime \triangleq [{{}^{C}\mathbf{n}_{l}^\prime}^\top, {{}^{C}\mathbf{v}_{l}^\prime}^\top]^\top$ represent the same 3D line in the camera PLK coordinate, based on the properties of PLK line as shown in Fig.~\ref{fig:perpendicular}, we can construct two constraints as:
\begin{equation}
    \begin{aligned}
        {}^{C}\mathbf{n}_{l}^{\prime} \cdot {}^{C}\mathbf{v}_{l} &= 0 \qquad\text{co-perpendicular constraint}, \\
        {}^{C}\mathbf{n}_{l}^{\prime} \times {}^{C}\mathbf{n}_{l} &= 0 \qquad\text{co-parallel constraint}.
        \label{eq:contraints}
    \end{aligned}
\end{equation}
% A 3D-to-2D line constraint can be expressed as:
% \begin{equation}
%     \begin{aligned}
%         \left[\begin{array}{c}l_1^\prime \\ l_2^\prime \\ l_3^\prime\end{array}\right] = 
%         \left[\begin{array}{cc}
%         \mathbf{K} & \mathbf{0}_3
%         \end{array}\right]
%         \left[\begin{array}{cc}
%     		{ }_{L}^{C}\mathbf{R} & \lfloor { }^{C}\mathbf{P}_{L}\times\rfloor { }_{L}^{C}\mathbf{R} \vspace*{2pt}\\
%     		\mathbf{0}_3 & { }_{L}^{C}\mathbf{R}
%     	\end{array}\right] \left[\begin{array}{c}{}^{L} \mathbf{n}_{l} \\ { }^{L} \mathbf{v}_{e}\end{array}\right] = 
%         \left[\begin{array}{cc}
%         \mathbf{K} & \mathbf{0}_3
%         \end{array}\right]\left[\begin{array}{c} { }_{L}^{C}\mathbf{R} {}^{L} \mathbf{n}_{l} + \lfloor { }^{C}\mathbf{P}_{L}\times\rfloor { }_{L}^{C}\mathbf{R} {}^{L} \mathbf{v}_{e} \vspace*{2pt}\\ { }_{L}^{C}\mathbf{R} { }^{L} \mathbf{v}_{e}\end{array}\right],
%         \label{eq:line_projection_loss}
%     \end{aligned}
% \end{equation}
% where $l_1^\prime=v_e-v_s$, $l_2^\prime=u_s-u_e$, and $l_3^\prime=u_ev_s-u_sv_e$ are 2D line parameters calculated from two endpoints $[u_s, v_s]$ and $[u_e, v_e]$. 
From \eqref{eq:line_transformation2} and \eqref{eq:contraints}, we find that the co-perpendicular constraint is only related to ${}_{L}^{C}\mathbf{R}$, which can be written as:
\begin{equation}
    \begin{aligned}
        \left[\begin{array}{ccc}l_1^\prime & l_2^\prime & l_3^\prime\end{array}\right]
        \left(\mathbf{K}^{-1}\right)^T {}_{L}^{C}\mathbf{R} { }^{L}\mathbf{v}_{l} &= 0.
        \label{eq:method2_constrant}
    \end{aligned}
\end{equation}

To solve ${}_{L}^{C}\mathbf{R}$, we define the residual of each correspondence as:
\begin{equation}
    r^\prime = \left[\begin{array}{ccc}l_1^\prime & l_2^\prime & l_3^\prime\end{array}\right]
        \left(\mathbf{K}^{-1}\right)^T {}_{L}^{C}\mathbf{R} { }^{L}\mathbf{v}_{l},
    \label{eq:rot_residual}
\end{equation}
and ${}_{L}^{C}\mathbf{R}$ can be solved by: 
\begin{equation}
    \underset{{}_{L}^{C}\mathbf{\tilde{\theta}}}{\text{minimize}} \sum_{n=1}^{N} ({r^\prime_n})^2.
    \label{eq:theta_minimal}
\end{equation}
To efficiently use non-linear optimizers, the analytical derivation of Jacobians can be expressed as:
\begin{equation}
    \begin{aligned}
        \frac{\partial {r^\prime}^2}{\partial {}_{L}^{C}\mathbf{\tilde{\theta}}} = 2r^\prime (- \left[\begin{array}{ccc}l_1^\prime & l_2^\prime & l_3^\prime\end{array}\right]
        \left(\mathbf{K}^{-1}\right)^T \lfloor{ }^{L}\mathbf{v}_{l}\times\rfloor)
        \label{eq:triangulation_Jacobians}.
    \end{aligned}
\end{equation}

Once ${}_{L}^{C}\mathbf{R}$ is determined, the ${ }^{C}\mathbf{P}_{L}$ can be estimated using the second constraint in \eqref{eq:contraints} as:
\begin{equation}
    \begin{aligned}
        \left(\mathbf{K}^{-1} \left[\begin{array}{c}l_1^\prime \\ l_2^\prime \\ l_3^\prime\end{array}\right]\right) \times 
    	(&{}_{L}^{C}\mathbf{R} {}^{L} \mathbf{n}_{l} + \left\lfloor { }^{C}\mathbf{P}_{L}\times\right\rfloor { }_{L}^{C}\mathbf{R} {}^{L} \mathbf{v}_{l}) = \mathbf{0},\\
        \Rightarrow \left(\mathbf{K}^{-1} \left[\begin{array}{c}l_1^\prime \\ l_2^\prime \\ l_3^\prime\end{array}\right]\right) \times
    	(&{}_{L}^{C}\mathbf{R} {}^{L} \mathbf{n}_{l} - \left\lfloor {}_{L}^{C}\mathbf{R} {}^{L} \mathbf{v}_{l} \times\right\rfloor { }^{C}\mathbf{P}_{L}) = \mathbf{0}, \\
        \Rightarrow  \left\lfloor\left(\mathbf{K}^{-1} \left[\begin{array}{c}l_1^\prime \\ l_2^\prime \\ l_3^\prime\end{array}\right]\right) \times\right\rfloor &\left\lfloor {}_{L}^{C}\mathbf{R} {}^{L} \mathbf{v}_{l} \times \right\rfloor { }^{C}\mathbf{P}_{L} = \\
        &\left\lfloor\left(\mathbf{K}^{-1} \left[\begin{array}{c}l_1^\prime \\ l_2^\prime \\ l_3^\prime\end{array}\right]\right) \times\right\rfloor {}_{L}^{C}\mathbf{R} {}^{L} \mathbf{n}_{l},
        \label{eq:constraint_t}
    \end{aligned}
\end{equation}
where ${}^{C}\mathbf{P}_{L}$ is the only unknown. The optimal solution can be found as a linear least square problem using singular value decomposition (SVD).

\section{SIMULATIONS AND DEGENERATE ANALYSIS}\label{sec:Simulations}
In this section, we analyze the degenerate cases for the two proposed algorithms and verify the solution using Monte Carlo simulations. For the first algorithm, each line pair provides two constraints in \eqref{eq:line_residual}, requiring three lines to solve for six unknowns. Similarly, for the second algorithm, both \eqref{eq:rot_residual} and \eqref{eq:constraint_t} offer one constraint each, so three lines are needed to solve rotation and translation independently.

According to the definition of PLK line in \eqref{eq:pluker_line}, the vectors ${}^{L}\mathbf{v}_{e}$ of three lines are identical when they are parallel in 3D. Since ${}^{L}\mathbf{v}_{l} = {}^{L}\mathbf{v}_{e}$, the only variable in \eqref{eq:method2_constrant} is $\mathbf{l}^\prime = \left[\begin{array}{ccc}l_1^\prime & l_2^\prime & l_3^\prime\end{array}\right]$. However, if we rotate the image plane around the axis perpendicular to the red line plane as in Fig.~\ref{fig:perpendicular}, the line measurement equation remains the same, and the system loses at least one rotational constraint along this direction. On the other hand, the term $\left\lfloor {}_{L}^{C}\mathbf{R} {}^{L} \mathbf{v}_{l} \times \right\rfloor$ in \eqref{eq:constraint_t} remains constant for all three lines. This makes the translation ${}^{C}\mathbf{P}_{L}$ only dependent on ${}^{L}\mathbf{n}_{l}$. As a result, the system loses at least one translation constraint along the direction of the lines and makes it unsolvable.

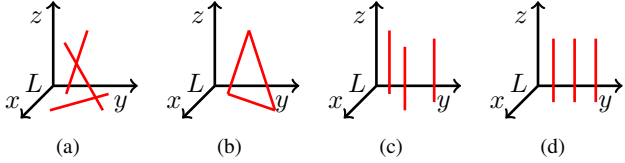
\begin{figure}[t]
    \centering
    \subfigure[]{\label{fig:normal}\begin{tikzpicture}[line width=1pt,scale=0.28]
\draw[->] (0,0,0) -- (0,0,4) node[anchor=south, xshift=-3]{$x$};
\draw[->] (0,0,0) -- (4,0,0) node[anchor=north east]{$y$};
\draw[->] (0,0,0) -- (0,4,0) node[anchor=north east]{$z$};
\node at (0,0,0) [anchor=south east, yshift=-6] {$L$};

\draw[red] (3,0,1) -- (1,0,3);
\draw[red] (1,0,1) -- (2,3,1);
\draw[red] (1.5,3,2.5) -- (3.5,0,3);
\end{tikzpicture}}
    \subfigure[]{\label{fig:coplanar}\begin{tikzpicture}[line width=1pt,scale=0.28]
\draw[->] (0,0,0) -- (0,0,4) node[anchor=south, xshift=-3]{$x$};
\draw[->] (0,0,0) -- (4,0,0) node[anchor=north east]{$y$};
\draw[->] (0,0,0) -- (0,4,0) node[anchor=north east]{$z$};
\node at (0,0,0) [anchor=south east, yshift=-6] {$L$};

\draw[red] (1,0,1) -- (4,0,3);
\draw[red] (1,0,1) -- (2,3,1);
\draw[red] (2,3,1) -- (4,0,3);
\end{tikzpicture}}
    \subfigure[]{\label{fig:parallel}\begin{tikzpicture}[line width=1pt,scale=0.28]
\draw[->] (0,0,0) -- (0,0,4) node[anchor=south, xshift=-3]{$x$};
\draw[->] (0,0,0) -- (4,0,0) node[anchor=north east]{$y$};
\draw[->] (0,0,0) -- (0,4,0) node[anchor=north east]{$z$};
\node at (0,0,0) [anchor=south east, yshift=-6] {$L$};

\draw[red] (1,0,1) -- (1,3,1);
\draw[red] (3.5,0,2) -- (3.5,3,2);
\draw[red] (2.5,0,3) -- (2.5,3,3);
\end{tikzpicture}}
    \subfigure[]{\label{fig:coplanar_and_parallel}\begin{tikzpicture}[line width=1pt,scale=0.28]
\draw[->] (0,0,0) -- (0,0,4) node[anchor=south, xshift=-3]{$x$};
\draw[->] (0,0,0) -- (4,0,0) node[anchor=north east]{$y$};
\draw[->] (0,0,0) -- (0,4,0) node[anchor=north east]{$z$};
\node at (0,0,0) [anchor=south east, yshift=-6] {$L$};

\draw[red] (1.5,0,2) -- (1.5,3,2);
\draw[red] (2.5,0,2) -- (2.5,3,2);
\draw[red] (3.5,0,2) -- (3.5,3,2);
\end{tikzpicture}}
    \caption{(a) Normal, (b) Coplanar, (c) Parallel, (d) Coplanar and parallel.}
    \vspace*{-15pt}
    \label{fig:observability}
\end{figure}
In Monte Carlo simulations, we explore four scenarios involving three line pairs as illustrated in Fig.~\ref{fig:observability}: (a) three nonparallel, non-coplanar lines, (b) three coplanar lines, (c) three parallel lines, and (d) three coplanar and parallel lines. Specifically, we generate three lines in LiDAR coordinate and transform them into the camera frame with extrinsic groundtruth. Then, we project the lines to the image frame and introduce projection noise with one-sigma variance. The initial guess is obtained by applying a known translation and rotation to the groundtruth. Finally, we conduct 10 Monte Carlo simulations with different random noises using two algorithms in four scenarios. 

The errors are calculated using the $L_2$ norm, and the simulation results are shown in TABLE~\ref{table:sim_results}. Notably, the orientation and translation errors for parallel line cases (the last two rows) are significantly higher than others. \textit{This indicates that three nonparallel lines are the minimum requirement to estimate extrinsic parameters for both algorithms.} Additionally, for scenarios (a) and (b), it is clear that PLK-Calib results in lower errors, particularly in the translation. As discussed in Sec.~\ref{sec:Introduction}, a small orientation error can lead to a large translation error. PLK-Calib imposes stronger constraints on the orientation, leading to a more accurate estimate. In contrast, the first algorithm combines orientation and translation errors into one loss function, causing them to restrict each other and reach a local minimum for both.
\begin{figure}[t]
	\centering
    \includegraphics[width=\columnwidth]{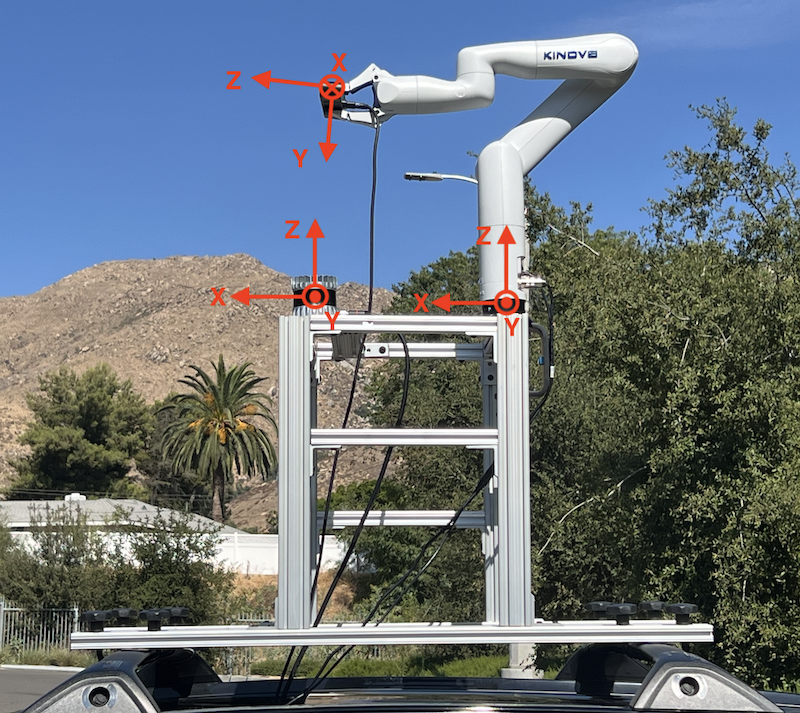}
    \vspace*{-15pt}
	\caption{Hardware setup. An Ouster OS1-128 LiDAR and a Kinova Gen3 lite mechanical arm are mounted on a sensor tower, and placed on the roof of a vehicle. The end effector holds a Zed 2i camera, and can navigate to different poses in 3D. The red arrows denote the coordinates of camera, LiDAR, and base frames.}
	\label{fig:hardware}
    \vspace*{-15pt}
\end{figure}
\begin{table}[t]
	\centering
	\caption{The $L_2$ norm of the orientation / translation (degrees / meters) errors of two methods using four different scenarios in Fig.~\ref{fig:observability}.}
	\begin{tabular}{ c | c c }
		\toprule
         & \textbf{Method I} & \textbf{PLK-Calib (Method II)}\\
		\midrule
        \textbf{(a)} & 0.003$\pm$0.003 / 0.111$\pm$0.072 & 0.002$\pm$0.001 / 0.040$\pm$0.047\\
		\textbf{(b)} & 0.004$\pm$0.003/ 0.129$\pm$0.164 & 0.004$\pm$0.005 / 0.066$\pm$0.055\\ 
		\textbf{(c)} & 0.073$\pm$0.068 / 98$\pm$608 & 0.165$\pm$0.000 / 905$\pm$1544\\
        \textbf{(d)} & 0.058$\pm$0.097 / 349$\pm$2089 & 0.164$\pm$0.000 / 1530$\pm$2163\\
		\bottomrule
	\end{tabular}
    \vspace*{-10pt}
	\label{table:sim_results}
\end{table}

\begin{figure*}[t]
    \centering
    \includegraphics[width=0.19\textwidth]{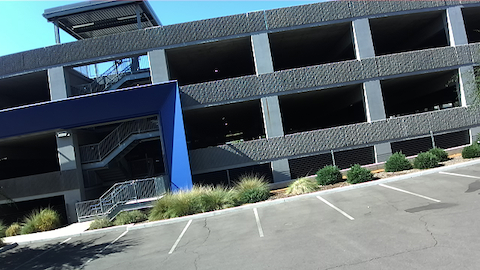}
    \includegraphics[width=0.19\textwidth]{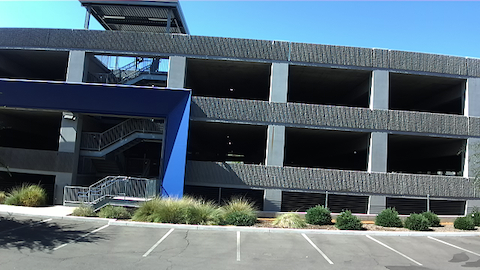}
    \includegraphics[width=0.19\textwidth]{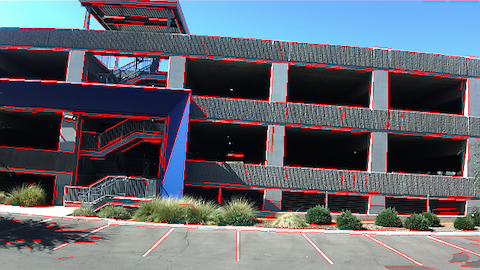}
    \includegraphics[width=0.19\textwidth]{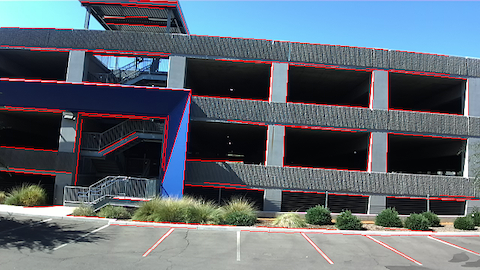}
    \includegraphics[width=0.19\textwidth]{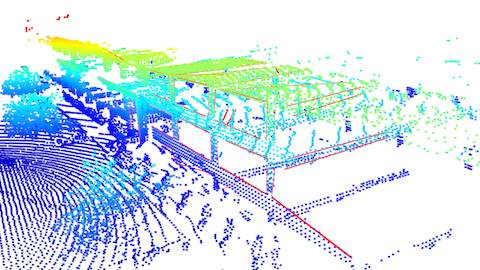}\\
    \vspace{0.1cm}
    \includegraphics[width=0.19\textwidth]{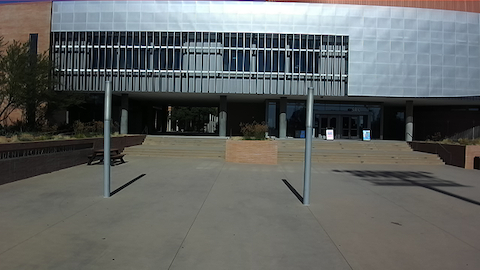}
    \includegraphics[width=0.19\textwidth]{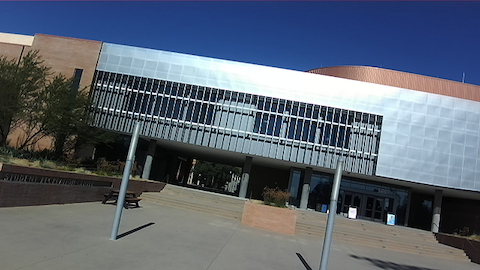}
    \includegraphics[width=0.19\textwidth]{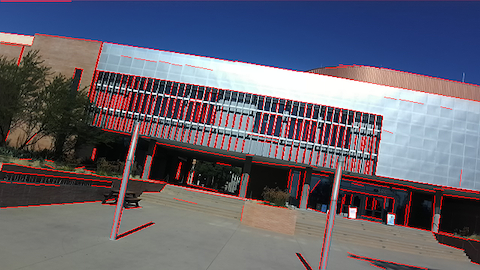}
    \includegraphics[width=0.19\textwidth]{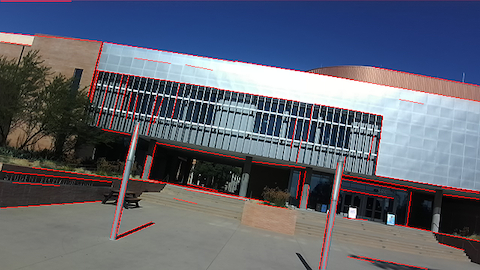}
    \includegraphics[width=0.19\textwidth]{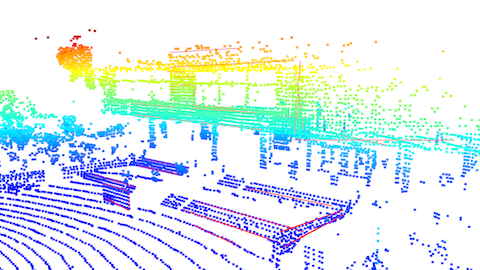}\\
    \vspace{0.1cm}
    \includegraphics[width=0.19\textwidth]{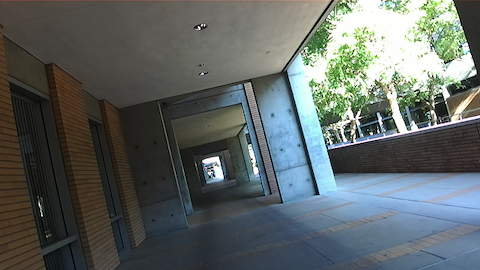}
    \includegraphics[width=0.19\textwidth]{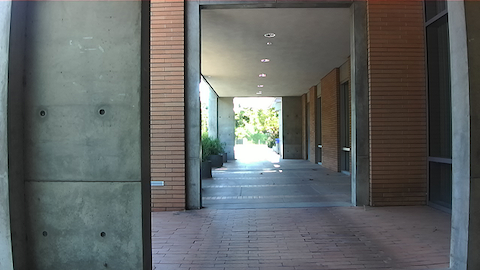}
    \includegraphics[width=0.19\textwidth]{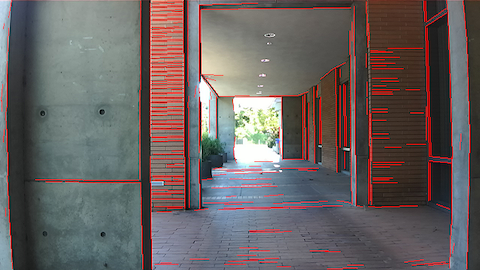}
    \includegraphics[width=0.19\textwidth]{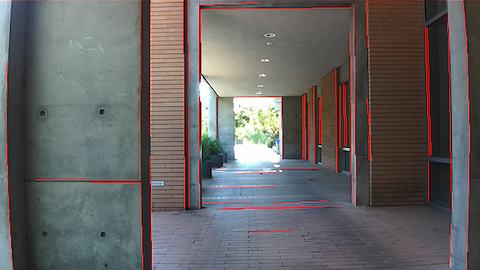}
    \includegraphics[width=0.19\textwidth]{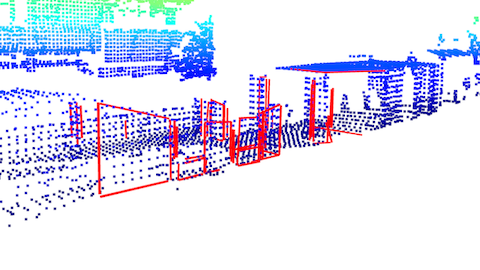}\\
    % \vspace*{-5pt}
    \caption{Dataset examples and experiment results across three different scenarios: parking lot, gym, and yard. The first two columns show dataset examples captured at various poses. The third and fourth columns display line detection results from images before and after applying combination and removal strategies. The final column illustrates line detection from the point clouds.}
    \label{fig:dataset}
    \vspace*{-15pt}
\end{figure*}
\section{EXPERIMENTS}\label{sec:Experiments}
In this section, we evaluate the proposed algorithm using a novel LC calibration dataset, collected by an Ouster OS1-128 LiDAR, a Zed 2i camera, and a Kinova Gen3 lite mechanical arm as shown in Fig.~\ref{fig:hardware}. Unlike all previous datasets, we do not fix the relative pose between the LiDAR and camera, allowing for the collection of multiple LC pairs with different extrinsic parameters within each scenario.

% \subsection{Dataset Literature}\label{sec:DatasetLiterature}
% Previous works on LC calibration and datasets can be grouped into three main categories: simulated dataset, public dataset, and customized dataset. Simulated datasets are generated using rendering software like Carla \cite{carla} or Gazebo \cite{gazebo}, where measurements and ground truth are generated from a controlled environment \cite{Manivasagam2020, Fang2021}. The main advantage of simulated datasets is that they allow for easy verification of the proposed algorithm's precision, as the ground truth for both point clouds and images is known. However, the simulated environment is not complex as the real-world, and cannot accurately reflect real-world accuracy. Public datasets, such as KITTI \cite{Geiger2012} and nuScenes \cite{Caesar2020}, are widely used for evaluation \cite{Huang2024, Petek2024, Zhaotong2023, Iyer2018, Lv2021, Schneider2017} because they provide groundtruth of extrinsic parameters, and make it easy to compare with other works. Recently, customized datasets have been introduced to meet specific hardware or environmental requirements \cite{Yuan2021, Huang2024_2}. 

% However, in all of these datasets, the relative transformation between the LiDAR and camera is fixed during recording, and they only consider one specific extrinsic parameter in the whole dataset. 

\subsection{Dataset Collection}\label{sec:DatasetCollection}
We collect a novel dataset in three different scenarios: parking lot, gym, and yard, located on the main campus of the University of California, Riverside. In each scenario, we navigate the mechanical arm to five different 3D poses, and record 40 LC pairs at each pose as shown in the first two columns of Fig.~\ref{fig:dataset}. Specifically, an Ouster OS1-128 LiDAR and the base of a Kinova Gen3 lite mechanical arm are mounted on a sensor tower, while a Zed 2i camera is attached to the arm's end effector as shown in Fig.~\ref{fig:hardware}. When launching the LiDAR Robot Operating System (ROS) \cite{Zhang2019} driver, we enable a high vertical resolution, resulting in $128 \times 2048 = 262144$ points in each LiDAR scan.

The relative pose between the end effector frame \{$E$\} and the base frame \{$B$\} is directly measured by the arm's encoder, and the repeatability can achieve $0.15$ mm. Besides, the relative pose between the camera frame and end effector frame, and between the base frame and LiDAR frame are measured manually. Consequently, the extrinsic groundtruth between LiDAR and camera can be expressed as: $\mathbf{T}_{L}^{C} = \mathbf{T}_{E}^{C} \, \mathbf{T}_{B}^{E} \, \mathbf{T}_{L}^{B}$, achieving centimeter-level accuracy. \textit{We avoided using the motion capture system for groundtruth, as marker placement at the LiDAR center and camera optical center is impossible. Combined with motion capture errors, the overall accuracy becomes uncontrollable.} Finally, the camera intrinsic parameters are calibrated using Kalibr \cite{Rehder2016}, and all collected LC pairs, along with their corresponding groundtruth, are available online\footnote{\url{https://drive.google.com/drive/folders/1pMGvzuJ0wV0cvyzI0R5IDbXbZSWsjOc1?usp=sharing}}.

\begin{figure}[t]
    \centering
    \includegraphics[width=\columnwidth]{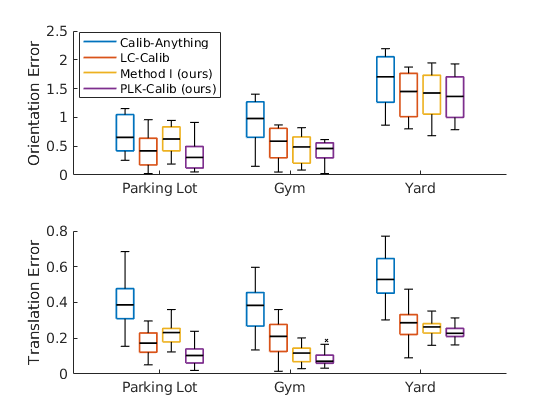}
    \vspace*{-15pt}
    \caption{The boxplot of $L_2$ norm of the orientation / translation (degrees / meters) errors using four algorithms in three different scenarios.}
    \label{fig:exp1}
    \vspace*{-15pt}
\end{figure}
\subsection{Evaluation}\label{sec:Evaluation}
% \begin{figure}[h]
%     \centering
%     \subfigure[]{\label{fig:line_combine}\input{Figures/line_combine.tikz}}
%     \subfigure[]{\label{fig:combine_example}\input{Figures/line_combine.tikz}}
%     \caption{(a) Normal (b) Coplanar}
%     % \vspace*{-20pt}
%     \label{fig:line_hander}
% \end{figure}

To extract line features from images, we use Line Segment Detector (LSD) \cite{Grompone2010, Zhang2025nerf} to identify two endpoints of each line. We then combine two lines if they meet the following criteria: (1) the distance from any endpoint on the first line to any endpoint on the second line is less than 5 pixels, mathematically expressed as $\min(\|\mathbf{x}_{1, s}-\mathbf{x}_{2, s}\|_2, \|\mathbf{x}_{1, s}-\mathbf{x}_{2, e}\|_2, \|\mathbf{x}_{1, e}-\mathbf{x}_{2, s}\|_2, \|\mathbf{x}_{1, e}-\mathbf{x}_{2, e}\|_2) < 5$, and (2) the angle difference between the two lines is less than $2$ degrees. Besides, we discard any line if its length is shorter than $20$ pixels. The detected lines before and after applying these merging and filtering strategies are shown in the third and fourth columns of Fig.~\ref{fig:dataset}, respectively. For line detection from an unorganized point clouds, we adopt the fast line extraction method described in \cite{Xiaohu2019} and the final detected 3D lines are shown in the last column of Fig.~\ref{fig:dataset}.

We compare the performance of our proposed algorithms with two open-source LC calibration methods: LC-Calib \cite{Koide2023} and Calib-Anything \cite{Zhaotong2023}. These two methods are selected because they are recent single-shot, target-less approaches. LC-Calib extracts point features from both images and point clouds, matching them based on the normalized information distance. Calib-Anything is a large vision model (LVM)-based method that utilizes Segment Anything \cite{Kirillov2023}. To ensure a fair comparison, we manually set the same initial guess and camera intrinsic parameters for both of these methods and ours. For the initial guess, we introduced a $5$-degree orientation error around each axis and a $0.5$-meter translation error along each axis relative to the groundtruth. In each scenario, data from $5$ different relative poses were selected, and $10$ LC pairs were selected at each pose, resulting in a total of $50$ pairs for each box plotted in Fig.~\ref{fig:exp1}. It is evident that PLK-Calib achieves similar performance to LC-Calib in terms of orientation estimates and surpasses all other algorithms in translation accuracy. Besides, Calib-Anything performs poorly in the yard scenario due to the limited number of segments detected.

% Notably, both LC-Calib and Calib-Anything converge to poor results when provided with a larger initial guess, such as a $2$-meter offset in translation. To evaluate the robustness of our algorithms under a bad initial guess, we conducted two additional experiments: (1) only translation offset along the LiDAR y-axis (horizontal offset) is added, and only the translation components are estimated while keeping the orientation using the groundtruth; and (2) only orientation offset around LiDAR z-axis (horizontal rotation) is introduced, and only the orientation components are estimated while keeping the translation parameters. We utilize PLK-Calib and the simplest gym scenario for this evaluation, and only use the top-$5$ matched LC pairs. The results are shown in Fig.~\ref{fig:exp2}, and indicate that PLK-Calib performs well even with a $5$-meter translation along the x-axis or a $25$-degree orientation around the z-axis in the LiDAR frame. The instances where both translation and orientation errors diverge to infinity are due to fewer than three non-parallel lines being projected into the image frame or incorrect line matching after extending a translation or orientation offset.

% \begin{figure}[H]
%     \centering
%     \includegraphics[width=\columnwidth]{Figures/exp2.png}
%     \vspace*{-15pt}
%     \caption{The $L_2$ norm of the translation / orientation (meters / degrees) errors after adding different translation and rotation offsets.}
%     \label{fig:exp2}
%     % \vspace*{-15pt}
% \end{figure}

\section{CONCLUSIONS}\label{sec:Conclusions}
In this paper, we proposed a local refinement algorithm for LC extrinsic calibration. Leveraging the co-perpendicular and co-parallel constraints of PLK lines, we decouple the rotation and translation into two constraints, resulting in more accurate estimates for both. Besides, we collected an LC calibration dataset and compared the accuracy of our algorithm with two baselines. The results demonstrate that PLK-Calib outperforms the baselines across all scenarios.

\section{ACKNOWLEDGEMENT}\label{sec:Acknowledgement}
We would like to thank the Autonomous Robots and Control Systems (ARCS) Laboratory and the Collaborative Intelligence Systems Laboratory (CISL) at the University of California, Riverside, for their generous hardware support.
%%%%%%%%%%%%%%%%%%%%%%%%%%%%%%%%%%%%%%%%%%%%%%%%%%%%%%%%%%%%%%%%%%%%%%%%%%%%%%%%
% \clearpage\newpage
% \balance
\bibliography{references}
\bibliographystyle{IEEEtran}

\end{document}